\documentclass[letterpaper, 10 pt, conference]{ieeeconf}  % Comment this line out if you need a4paper
\IEEEoverridecommandlockouts                              % This command is only needed if 
\usepackage[utf8]{inputenc}
\usepackage{graphicx} % for pdf, bitmapped graphics files
\usepackage{amsmath} % assumes amsmath package installed
\usepackage{amssymb}  % assumes amsmath package installed
\usepackage{mathtools}
\usepackage{accents}
\usepackage{booktabs}
\usepackage{multirow}
\usepackage[mode=buildnew]{standalone}
\usepackage{tikzscale}
\usepackage{subfig}
\usepackage{url}
\usepackage{xcolor}
\usepackage[hidelinks]{hyperref}
\usepackage[colorinlistoftodos]{todonotes} % ,disable
\usepackage{regexpatch}
\usepackage{siunitx}
\makeatletter
\xpatchcmd{\@todo}{\setkeys{todonotes}{#1}}{\setkeys{todonotes}{inline,#1}}{}{}
\makeatother

\let\norm\undefined % <-- "Undefine" \norm
\DeclarePairedDelimiter\norm{\lVert}{\rVert}

\newlength{\dhatheight}

\newboolean{long_version}
\setboolean{long_version}{true} % variable to switch between short (IV) and long (arXiv) version

\ifthenelse{\boolean{long_version}}
{%
	\title{\LARGE \bf GAN- vs. JPEG2000 Image Compression\\for Distributed Automotive Perception:\\Higher Peak SNR Does Not Mean Better Semantic Segmentation} % arXiv
}
{%
	\title{\LARGE \bf On Low-Bitrate Image Compression\\for Distributed Automotive Perception:\\Higher Peak SNR Does Not Mean Better Semantic Segmentation} % original
}

\author{Jonas Löhdefink$^{1}$, Andreas Bär$^{1}$, Nico M. Schmidt$^{2}$, Fabian Hüger$^{2}$, Peter Schlicht$^{2}$ and Tim Fingscheidt$^{1}$%
\thanks{$^{1}$Jonas Löhdefink, Andreas Bär and Tim Fingscheidt are with the Institute for Communications Technology,
	Technische Universität Braunschweig, Schleinitzstr. 22, 38106 Braunschweig, Germany
	{\tt\footnotesize \{j.loehdefink, andreas.baer, t.fingscheidt\}@tu-bs.de}}%
\thanks{$^{2}$Nico Maurice Schmidt, Fabian Hüger and Peter Schlicht are with Volkswagen Group Research -- Automated Driving, Berliner Ring 2, 38440 Wolfsburg, Germany
	{\tt\footnotesize \{nico.maurice.schmidt, fabian.hueger, peter.schlicht\}@volkswagen.de}}%
}

\begin{document}

\maketitle
\thispagestyle{empty}
\pagestyle{empty}

\begin{abstract}
The high amount of sensors required for autonomous driving poses enormous challenges on the capacity of automotive bus systems.
There is a need to understand tradeoffs between bitrate and perception performance.
In this paper, we compare the image compression standards JPEG, JPEG2000, and WebP to a modern encoder/decoder image compression approach based on generative adversarial networks (GANs).
We evaluate both the pure compression performance using typical metrics such as peak signal-to-noise ratio (PSNR), structural similarity (SSIM) and others, but also the performance of a subsequent perception function, namely a semantic segmentation (characterized by the mean intersection over union (mIoU) measure).
Not surprisingly, for all investigated compression methods, a higher bitrate means better results in all investigated quality metrics.
Interestingly, however, we show that the semantic segmentation mIoU of the GAN autoencoder in the highly relevant low-bitrate regime (at \SI{0.0625}{bit\per pixel}) is better by \SI{3.9}{\percent} absolute than JPEG2000, although the latter still is considerably better in terms of PSNR (\SI{5.91}{\decibel} difference).
% retrain semantic segmentation on compressed image
This effect can greatly be enlarged by training the semantic segmentation model with images originating from the decoder, so that the mIoU using the segmentation model trained by GAN reconstructions exceeds the use of the model trained with original images by almost \SI{20}{\percent} absolute.
We conclude that distributed perception in future autonomous driving will most probably \textit{not} provide a solution to the automotive bus capacity bottleneck by using standard compression schemes such as JPEG2000, but requires modern coding approaches, with the GAN encoder/decoder method being a promising candidate.
\end{abstract}

\section{Introduction}
\label{sec:introduction}
\begin{figure}[t!]
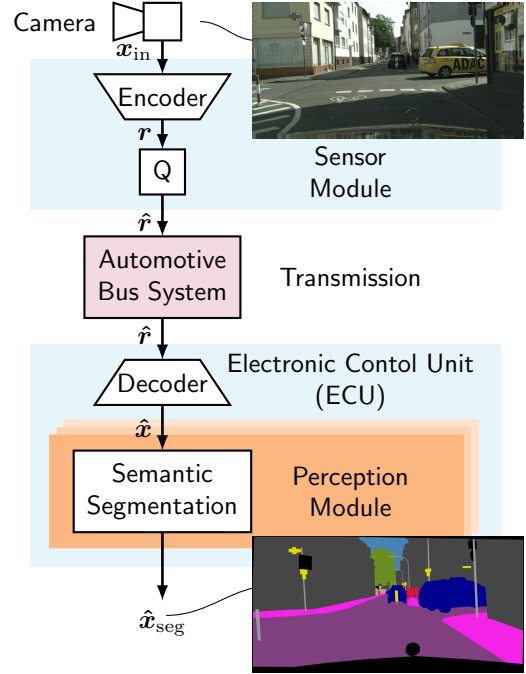

	\centering
	\includestandalone{inference_vert}
	\setlength{\belowcaptionskip}{-15pt}
	\caption{\textbf{Distributed automotive perception (inference setup)} with a camera, a sensor module for image compression including quantization (i.e., encoding), transmission over the automotive bus system, decoding, and a subsequent perception module performing semantic segmentation.}
	\label{fig:inference}
\end{figure}

% 1 Image Compression is necessary
In autonomous driving, perception incorporates many sensors to build an overall representation of the surrounding.
The processing and transmission of vast amounts of information from different sensors across the vehicle leads to a serious bottleneck.
Especially, the automotive bus system struggles to provide a sufficiently high bitrate, which is why we need to compress sensor outputs such as images.
An overview of the proposed system can be seen in Figure~\ref{fig:inference}, which shows inference by distributed perception in the vehicle.
The recorded data is encoded and quantized in the sensor module, decoded on the central electronic control unit (ECU), and afterwards processed by a semantic segmentation as an exemplary perception module.

% conventional and our framework
Conventional codecs for image compression such as JEPG~\cite{jpeg}, JPEG2000~\cite{jp2}, and WebP~\cite{webp} use loss functions optimized for an accurate reconstruction of local structures, e.g., the mean squared error (MSE).
However, this often results in perceptually poor-looking images once the bitrate decreases.
% 2 Higher PSNR does not mean better semantic segmentation
Besides the MSE, there exist several further image assessment metrics.
%image compression: conventional distortion metrics: PSNR, SSIM, MS-SSIM
The most traditional one is the peak signal-to-noise ratio (PSNR)~\cite{salomon2004data}, which is closely related to the MSE.
Furthermore, there are the structural similarity (SSIM)~\cite{wang2004image} and its successor, the multi-scale structural similarity (MS-SSIM)~\cite{wang2003multi}.
The latter two also take the spatial context of pixels into account.
Nevertheless, all these metrics concentrate on a comparison between the decoded image $\boldsymbol{\hat{x}}$ and the uncoded reference $\boldsymbol{x}$.
This is why we call them \textit{distortion metrics}.

% 4 Semantic Segmentation
%perception metric: mIoU
On the contrary, our goal is to preserve global structures.
A comparison on the pixel level is not reasonable, because the exact appearance of those pixels is irrelevant for many subsequent perception functions.
A typical perception function is semantic segmentation, which essentially means, assigning a discrete semantic class label to each pixel in the image.
The segmentation process results in a class mask containing the objects as discrete pixel values at their initial location.
The texture of the objects is lost, only their class is preserved (see example image at the bottom of Figure~\ref{fig:inference}).
The mean intersection over union (mIoU) is the common quality measure for segmentation and represents the amount of overlap between the ground truth and the predicted class mask.
Following~\cite{Isola2016} and \cite{wang2018high}, we use the mIoU as a metric for the preservation of global structure.
This method does not compare the network output to the input images but to the segmentation ground truth instead and can therefore be called a \textit{perception metric}.

% closure with thesis
The divergent focus of distortion and perception metrics results in a tradeoff to be made for compression methods~\cite{Blau2017}.
In this paper, we show that higher values for the distortion metrics does not necessarily imply preservation of semantic structures.

%GAN intro
In the recent past, a new approach called generative adversarial networks (GANs)~\cite{Goodfellow2014} has emerged in the field of generative modeling.
These networks seem promising with regards to preserving semantic structures.
A GAN is composed of a generator and a discriminator.
The task of the generator is the generation of realistic-looking images.
The task of the discriminator is to distinguish between fake images constructed by the generator and real images from the training distribution.
Both are trained in a simultaneous manner.
In case of a successful training process, an equilibrium is reached, i.e., neither of the two models can achieve better performance by only adjusting its own parameters.
% compression framework intro
We use a GAN adopted from~\cite{Agustsson2018} to generate reconstructions of an original image.
Our framework consists of a classical autoencoder, which implies an encoder/decoder structure with an internal compressed representation after encoding.

In this paper, we compare image compression standards to a modern GAN-based approach, using an autoencoder as generator.
In our experimental evaluation, the pure compression performance is measured by the typical image assessment metrics PSNR, SSIM, and MS-SSIM.
Additionally, we investigate the performance of a semantic segmentation as an exemplary subsequent perception function characterized by the mIoU, and compare this to semantic segmentation operating on uncompressed images.
Our results will show that the GAN-based approach is better suited for low-bitrate image compression than traditional codecs.
Apart from this, we will show that an end-to-end optimized scenario for both compression and semantic segmentation can reach even higher mIoU scores.

The rest of the paper is structured as follows:
Section~\ref{sec:related_work} gives an overview of related works.
Section~\ref{sec:image_compression_framework} presents the main compression framework.
The employed semantic segmentation network is outlined in Section~\ref{sec:semantic_segmentation}.
The experimental part in Section~\ref{sec:experiments} contains the comparison of compression approaches, along with a discussion.
Conclusions are drawn in Section~\ref{sec:conclusions}.

\section{Related Work}
\label{sec:related_work}
This work covers aspects from the fields of generative adversarial networks
\ifthenelse{\boolean{long_version}}
{%
	image compression, and semantic segmentation.
}
{%
	and image compression.
}

\subsection{Generative Adversarial Networks}
% theory: GAN, cGAN
In the last few years, generative adversarial networks (GANs), initially proposed in~\cite{Goodfellow2014}, gained much attention in the image processing domain.
To control the content of generated images, Mirza and Osindero proposed the conditional GAN~\cite{Mirza2014}, which uses conditioning variables as additional input for both the generator and the discriminator.
% general tips, DCGAN
Due to the simultaneous training process, GANs tend to be unstable.
To combat this effect, Radford et al.\ released several useful hints regarding the network designs~\cite{Radford2015}, including suggestions for architectures, normalizations, and activation functions.
% topologies: lsgan, ebgan, began, wgan
Since then, several investigations have been published to further improve the learning process~\cite{Mao2016, Zhao2016, Berthelot2017, Arjovsky2017, Kurach2018}.
They consider different loss functions used in the discriminator training~\cite{Kurach2018}, e.g., the least-square function~\cite{Mao2016} or the Wasserstein distance~\cite{Zhao2016} between generator distribution and target distribution.

GANs are used in a lot of diverse applications in which generative models are involved.
These include learning of data representations~\cite{Makhzani2015}, semantic segmentation~\cite{Luc2016}, teacher-student network compression~\cite{Belagiannis2018}, defending adversarial examples~\cite{Xiao2018, Samangouei2018, Ilyas2017}, and reinforcement learning~\cite{Ho2016}.
% automotive:
% generation of realistic simulations for training self-driving cars
The generation of training and validation material for autonomous driving systems is another use case of generative models.
Several works have been proposed to convert training data from a driving simulator or video games into real-world data samples~\cite{Shrivastava2016, Zhu2017, Li2018, Sankaranarayanan2017}.
This is referred to as a classical image-to-image transformation.
As will be discussed in the following, our major interest in GANs is for encoding and decoding of images for distributed perception in automotive applications.
Due to its stable training process and reasonable performance, we will build upon the least squares GAN (LS-GAN) from Mao et al.\ \cite{Mao2016}.

\subsection{Image Compression}
% general steps
The majority of compression methods is composed of similar stages, which include color transformations, block splitting, scalar quantization, and coding.
% JPEG
As one famous example, the JPEG standard~\cite{jpeg} uses a lossy \mbox{downsampling} of the chroma channels in addition and is based on the discrete cosine transform (DCT).
% JPEG2000
Opposed to this, the successor JPEG2000~\cite{jp2} is a wavelet-based compression method, with the major difference, that a discrete wavelet transform (DWT) is used to compute the coefficients to quantize.
% intraframe, interframe, WebP
As natural image data contains redundancy, it is possible to execute either inter-frame prediction (from temporally preceding images) or intra-frame prediction (from neighboring parts of the same image).
An example for the latter approach is the WebP codec~\cite{webp}, which is based on a DCT.

% autoencoder or recurrent neural network
The more recently proposed image compression methods are either based on autoencoders~\cite{Theis2017, Balle2016, Rippel2017} or recurrent neural networks, such as long short-term memory networks (LSTMs)~\cite{Toderici2015}.
The methods from~\cite{Rippel2017, Santurkar2017, Agustsson2018} also integrate an adversarial loss function to generate more \mbox{realistically} reconstructed images.
GAN-based approaches provide sharper reconstructions and higher compression rates.
In~\cite{Santurkar2017} only the decoder consists of the generator of a GAN, while in \cite{Agustsson2018} both the encoder and the decoder form the generator.
In this work, we will adopt the approach from~\cite{Agustsson2018} and will improve the performance on subsequent functions, since it provides a state-of-the-art LS-GAN combined with an autoencoder and a potentially low bitrate of the latent space.

\ifthenelse{\boolean{long_version}}
{%
	\subsection{Semantic Segmentation}
	% Introduction and FCNs
	In the context of image processing, semantic segmentation can be interpreted as a pixel-wise classification of an image.
	Current state-of-the-art models for semantic segmentation rely on the concept of fully convolutional networks (FCNs)~\cite{Long2015}, using a feature extractor pretrained on the ImageNet dataset~\cite{Russakovsky2015} and building a segmentation head on top.
	
	% Context module
	Residual networks (\texttt{ResNets})~\cite{He2016} are primarily chosen as feature extractors in many state-of-the-art models~\cite{Zhao2016a,Chen2017,Wu2016,Chen2018a,RotaBulo2018}.
	To aggregate more context while retaining the spatial resolution, these models make use of dilated convolution layers to enlarge the receptive field in deeper layers~\cite{Yu2016}.
	Further multi-scale context is addressed by building a pyramid pooling module on top of the feature extractor~\cite{Zhao2016a,Chen2017,Chen2018a,Chen2018,RotaBulo2018}.
	The models are designed in an encoder-decoder fashion to restore the original resolution.
	One simple approach is to use bilinear interpolation of the network prediction~\cite{Zhao2016a,Chen2017,Chen2018,RotaBulo2018}.
	Other approaches consider upsampling via transposed convolution~\cite{Wu2016} or the additional use of low-level features~\cite{Chen2018a,Ronneberger2015}.
	
	% Challenge: Memory and real-time
	Two closely related challenges in semantic segmentation are the limited GPU memory and computation time with regard to real-time applications.
	To address the memory problem,~\cite{RotaBulo2018} proposes in-place activated batch normalization (\texttt{InPlace-ABN}), a memory-efficient approach combining the leaky rectified linear unit (leaky ReLU) and batch normalization~\cite{Ioffe2015}.
	In~\cite{Chen2018a,Sandler2018} depthwise separable convolutions are used to reduce parameters and therefore computation cost and memory usage, while~\cite{Romera2018} proposes factorized convolution in combination with residual connections (as in~\cite{He2016}) to obtain a similar effect.
	In this work, we will focus on image compression rather than real-time semantic segmentation and therefore choose a semantic segmentation approach being similar to~\cite{RotaBulo2018}.
	
	\subsection{Quality Metrics}
	For the evaluation of compression approaches, quality measures for image restoration algorithms are used.
	We can separate them into distortion and perception metrics.
	The results of distortion measures and perception measures are not guaranteed to always correlate positively.
	This effect was already shown and proven in~\cite{Blau2017}.
	Typical distortion metrics are, e.g., the well-known PSNR, SSIM~\cite{wang2004image}, MS-SSIM~\cite{wang2003multi}.
	The most often used perception metric is the human opinion score, where a human has to decide, whether the image looks realistic or not.
	For GANs in general the inception score (IS)~\cite{salimans2016improved} and Frechet inception distance (FID)~\cite{Heusel2017} are often used~\cite{borji2018pros}.
	For our investigations we adopt PSNR, SSIM, and MS-SSIM for the basic compression scheme evaluations, and use the mIoU~\cite{Everingham2015} as quality measure for the semantic segmentation in form of a perception metric, thus selecting specifically suited metrics for different processing stages.
}
{%
	
}

\section{Image Compression Framework}
\label{sec:image_compression_framework}
\begin{figure}[t!]
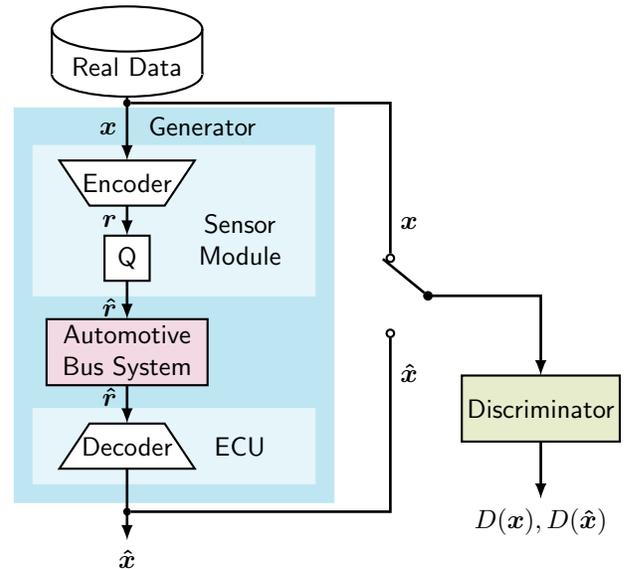

	\vspace*{0.15cm}
	\centering
	\includestandalone{framework_vert}
	\setlength{\belowcaptionskip}{-10pt}
	\caption{
		\textbf{Training setup} of our image compression framework based on~\cite{Agustsson2018}.
		The generator consists of an autoencoder with integrated quantization (Q) of the compressed representation, with decoding being performed on the electronic control unit (ECU).
		For test (i.e., inference), please see Figure~\ref{fig:inference}.}
	\label{fig:Base_framework}
	\vspace*{-0.15cm}
\end{figure}

This section deals with the general image encoding (compression) and decoding by means of a so-called autoencoder.
First we describe the framework as a whole and then we will go into details for each part of it.
\ifthenelse{\boolean{long_version}}
{%
	
	% Introduction of compression framework with GANs
	% Structure in inference/vehicle
	The framework for image compression is adopted from~\cite{Agustsson2018}, using primarily the network topologies.
	As \mbox{Agustsson} et al.\ did not publish any implementation details, the reimplementation from~\cite{JTan} was taken as a basis for our work.
	The training framework mainly consists of three neural networks as depicted in Figure~\ref{fig:Base_framework}, the encoder, the decoder and the discriminator.
	The compound of the first two networks constitutes the generator as the counterpart to the discriminator.
	This structure forms a GAN, where the generator is defined as an autoencoder.
	The output of the encoder is quantized and in our application transmitted over an automotive bus system before feeding the data to the decoder.
	Note that, within the scope of this work, this bus system is modeled as error-free transmission.
	During test (inference), only the generator of the GAN is used to compress the data and decode it after transmission (compare Figure~\ref{fig:inference} and~\ref{fig:Base_framework}).
	The encoder is located in the sensor module and compresses the image taken by the camera.
	The resulting bitstream is transmitted via the automotive bus system.
	On the receiver side, the decoder reconstructs the original image.
	In the following, all parts of the framework are described in detail, while the network architectures are specified in even more detail in~\cite{Agustsson2018}.
}
{%
	The framework for image compression is adopted from~\cite{Agustsson2018}, where the code reimplementation from~\cite{JTan} is the basis for our work.
	The training framework mainly consists of three neural networks as depicted in Figure~\ref{fig:Base_framework}, the encoder, the decoder and the discriminator.
}

\subsection{Encoder}
\label{sec:enc}
The task of the encoder is to produce a compact \mbox{representation} of the input image data.
The structure of the encoder consists of six convolutional blocks, each of these being built from a convolutional layer, an instance norm layer, and a ReLU activation function.
Downsampling is achieved by using a stride of $s=2$ from the 2nd up to the 5th convolutional block.
At the same time, the number of feature maps increases towards the latter convolutional blocks.
The last convolutional block yields $F\in\{2,4,8,10\}$ feature maps of $32 \times 64$ pixels (training) or $64 \times 128$ pixels (inference), altogether defining the bottleneck dimension.
The different latent space sizes result from different input image sizes in training and inference, see Section~\ref{sec:experiments:bitrate}.
The bottleneck dimension is actually a hyperparameter, which is used to control the size of the compressed image.

\subsection{Quantization}
\label{sec:quant}
The output of the encoder is quantized to obtain some significant image compression.
A simple approach would be to train the network without this quantization and use quantization only during inference.
Alternatively, we can integrate the quantization into the training, so the networks can adjust to the discrete input~\cite{bengio-arxiv13-condcomp}.
In Section~\ref{sec:experiments} we will compare the performance of both strategies.
As quantization produces discrete activations (grey values), its derivative equals zero almost everywhere.
Therefore, if quantization is integrated into the training, the learning process of the networks is hindered.
This problem can be circumvented by approximating the quantizer characteristic and using the derivative of this approximation.

There are several possibilities to approximate the quantizer characteristic, the easiest way being just using the identity function.
In this work, a softmax-based function is used to obtain a smooth approximation of quantizer output pixel $\hat{r}$ (see Figure~\ref{fig:inference})~\cite{JTan}:
\setlength{\abovedisplayskip}{5pt}
\setlength{\belowdisplayskip}{5pt}
\vspace{-0.1cm}
\begin{align}
\label{eq:est1}
\boldsymbol{e} &= \bigl|\boldsymbol{c} - \boldsymbol{r}\bigr|\\
\label{eq:est2}
\hat{r} &= \boldsymbol{c}^\intercal \cdot \frac{\exp(-\boldsymbol{e})}{\norm{\exp(-\boldsymbol{e})}_1},
\end{align}
where $\boldsymbol{r} = (r \cdots r)^\intercal$ is a column vector consisting of the quantizer input $r$ stacked $L$ times so that the dimension equals that of the column vector of the quantizer reconstruction levels $\boldsymbol{c} = (c_1 \cdots c_L)^\intercal$, with $()^\intercal$ being the transpose.
The resulting absolute error vector $\boldsymbol{e}$ is then subject to a softmax function (note the element-wise exponential function and division in~\eqref{eq:est2}) and afterwards multiplied by the transposed reconstruction levels row vector $\boldsymbol{c}^\intercal$ to obtain the approximated scalar quantizer output $\hat{r}$.
Note that $\norm{\cdot}_1$ is the L$_1$ norm, here effectively summing up all elements of its (positive) argument.

\subsection{Decoder}
\label{sec:dec}
The decoder acts as the counterpart to the encoder and reconstructs the original image.
The first part of the decoder consists of nine residual units, where convolutions and an identity function are processed in parallel.
This enables a deep network structure while maintaining a good learning behavior.
As the encoder performs a downsampling by a factor of $d$ in each dimension, the decoder analogously has to perform an upsampling by the same factor.
This is achieved by using subsequent transposed convolutions, which correspond to learnable upsampling functions.
The output of the decoder is the reconstructed image.

\begin{figure}[t!]
	\vspace*{0.1cm}
	\centering
	\includestandalone[scale=0.75]{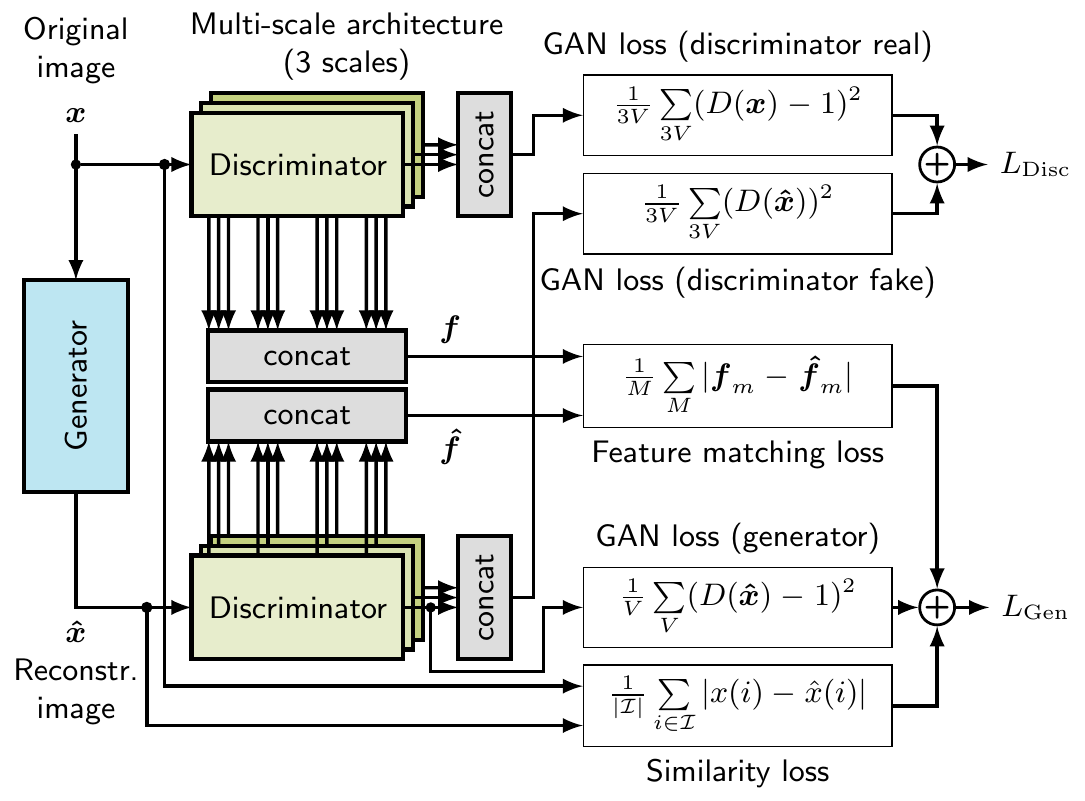}
	\setlength{\belowcaptionskip}{-10pt}
	\caption{
		\textbf{Training setup and losses} of the compression framework.
		The discriminator loss $L_{\textrm{Disc}}$ is the standard GAN loss, whereas the generator loss $L_{\textrm{Gen}}$ is composed of a GAN loss, a feature matching loss, and a similarity loss.
		With $\mathbb{G} = [-1,1]$ being the set of gray values and $i \in \mathcal{I}$ being the pixel index, the images $\boldsymbol{x}$, $\boldsymbol{\hat{x}} \in \mathbb{G}^{H \times W \times C}$ with image pixels $x(i), \hat{x}(i) \in \mathbb{G}$ consist of $|\mathcal{I}| = H \times W \times C$ pixels with the image height $H$,  the image width $W$, and $C = 3$ image channels.
	}
	\label{fig:losses}
	\vspace*{-0.1cm}
\end{figure}

\subsection{Discriminator, Training, and Loss Functions}
\label{sec:losses}
As the discriminator, a multi-scale \texttt{PatchGAN}~\cite{Isola2016} is used.
In this architecture, the input image is processed in multiple scales, here three scales are used.
The \texttt{PatchGAN} defines a small receptive field of size $v \times v$ instead of the whole image.
This enables the discriminator to focus on the classification of local structures.
The training protocol for both generator and discriminator uses three loss functions during the training, which are simply added up, as seen in Figure~\ref{fig:losses}.
First, the GAN loss is derived from the discriminator network, which tries to classify whether its input images are real or fake.
The outputs of the $3V$ different \texttt{PatchGAN} patches ($V$ patches in each scale) are evaluated afterwards for the real images as well as for the fake images.
Second, the similarity loss measures the MSE distortion on pixel level.
Finally, a feature matching loss is applied based on the assumption that the discriminator learns a representation of the data distribution.
Therefore, the discriminator is well-suited as a feature extractor.
This loss function compares the $M$ feature maps between the internal discriminator convolutions of the original and the reconstructed image.

\ifthenelse{\boolean{long_version}}
{%
	% Feeding of data, training procedure
	In the training process of GANs~\cite{Goodfellow2014}, first, the generator is trained, consuming a sample $\boldsymbol{x}$ which originates from the training set, resulting in a reconstructed image $\boldsymbol{\hat{x}}$ (Figure~\ref{fig:losses}).
	After the consecutive forward pass of the discriminator, the gradients of the generator loss $L_{\textrm{Gen}}$ are passed back through the discriminator and are then applied to the weights of the generator.
	Afterwards, in the discriminator training phase, another reconstruction is produced, which is used in combination with the label ``fake'', as well as an original image with the label ``real'' to train the discriminator (contributions to $L_{\textrm{Disc}}$).
}
{%
}

\subsection{Bitrate}
\label{sec:compression_rate}
The bitrate $R$ depends on the size of the latent space with respect to the original image size $H \times W$.
The downsampling factor $d$ represents the overall dimension reduction, which is defined by the number of convolutional layers $n = 4$ with a stride $s=2$, each introducing a downsampling by factor 2 in each of the image dimensions:
\begin{equation}
d = s^n = 2^4 = 16
\end{equation}
Apart from the downsampling factor $d^2$, also the image dimensions $H \times W$, the number of feature maps in the bottleneck $F$, and the number of scalar quantization levels $L$ are required for the computation of the latent space information $S$ in bit:
\begin{equation}
S = \frac{H}{d} \cdot \frac{W}{d} \cdot F \cdot \operatorname{ld}(L) = \frac{H \cdot W \cdot F \cdot \operatorname{ld}(L)}{d^2} \ \ \mathrm{[bit]}
\end{equation}
The bitrate $R$ in bit per pixel (bpp) can be computed by the latent space information $S$ and the original image dimensions $H \times W$ as follows:
\begin{equation}
\label{eq:bit_rate}
R = \frac{S}{H \cdot W} = \frac{F \cdot \operatorname{ld}(L)}{d^2} \ \ \mathrm{[bpp]}
\end{equation}

\section{Semantic Segmentation}
\label{sec:semantic_segmentation}
For the purpose of this work, we chose a semantic segmentation network which is commonly used for most perception tasks in automotive applications.
\ifthenelse{\boolean{long_version}}
{%
	This section briefly sketches the architecture and training as well as the evaluation of the underlying model for semantic segmentation.
	
	\subsection{Architecture}
	\label{sec:architecture}
	% Architecture description
	For the task of semantic segmentation we use a similiar approach as described in~\cite{RotaBulo2018} and use a pretrained feature extractor with a subsequent segmentation head trained by ourselves.
	We rebuild \texttt{DeepLabv3}~\cite{Chen2017} in \texttt{TensorFlow}~\cite{Abadi2016} and replace \texttt{ResNet50}~\cite{He2016} with a rebuilt and ImageNet-pretrained \texttt{ResNet38}~\cite{Wu2016} as the new feature extractor, with the difference to~\cite{RotaBulo2018} that we do not incorporate \texttt{InPlace-ABN}.
	
	Further, we made a few common modifications in semantic segmentation to \texttt{ResNet38}~\cite{Chen2017,Wu2016}.
	First, we removed the classification layer of \texttt{ResNet38} and connected the remaining network to the segmentation head of \texttt{DeepLabv3}.
	\mbox{Second}, to control the output stride, we decreased the stride of several convolutions from two to one in a bottom-up \mbox{fashion} and increased the dilation rate instead.
	Similar to ~\cite{Chen2017}, we refer to output stride as the overall downsampling factor of the network.
	To be more clear, consider a residual block consisting of several residual units, where inside the first residual unit a convolution with stride $s=2$ is performed.
	Here, we removed the striding operation and increased the dilation rate vertically and horizontally from one to two in all convolutions within the residual block.
	We did the same in the next residual block, but doubled the dilation rate from two to four instead.
	Third, in contrast to~\cite{Wu2016}, we do not incorporate dropout due to an observed worse performance.
	
	\subsection{Training}
	\label{sec:resnet38_training}
	For optimization we used the stochastic gradient descent (SGD) with momentum and a learning rate with polynomial decay as described in~\cite{Chen2017}.
	The input images are randomly resized $\left[ 0.5, 2.0\right]$, left-right flipped and cropped to the size 700x700.
	With this configuration and our reimplementations we could manage the batch size \mbox{$B_1=4$} for an output stride $s_1=16$ and batch size $B_2=2$ for an output stride $s_2=8$ on an \texttt{Nvidia Geforce GTX 1080 Ti} with \SI{11}{GB} RAM.
	
	The training itself can be described as a two-stage process.
	In the first stage, we set the output stride to $s_1=16$ and train the network parameters, including the batch normalization parameters, for 90,000 iterations with a batch size \mbox{$B_1=4$} and an initial learning rate $\eta_{\mathrm{1}}=0.001$.
	In some cases we had to further reduce the initial learning rate of the polynomial learning rate schedule in the second stage due to instability problems.
	In the second stage, we employ output stride \mbox{$s_2=8$}, freeze the batch normalization parameters, and fine-tune for additional 120,000 iterations with a reduced batch size \mbox{$B_2=2$} and a reduced initial learning rate \mbox{$\eta_{\mathrm{2}}=0.0005$}.
}
{%
	For the task of semantic segmentation we use a pretrained feature extractor with a subsequent segmentation head trained by ourselves.
	We rebuild \texttt{DeepLabv3}~\cite{Chen2017} and replace \texttt{ResNet50}~\cite{He2016} with a rebuilt and ImageNet-pretrained \texttt{ResNet38}~\cite{Wu2016} as the new feature extractor.
}

\section{Experiments and Discussion}
\label{sec:experiments}
% Introduction of the experimental part
The experiments in this paper are split into two parts.
The first part deals with varying the bitrate by adjusting the size of the bottleneck $H \times W \times F$ and the number of reconstruction levels $L$, to explore the tradeoff w.r.t. the reconstructed image quality (Section~\ref{sec:experiments:bitrate}).
The second part is about applying the semantic segmentation to the employed compression framework, first retraining the segmentation incorporating coded images and reporting the results on the validation set (Section~\ref{sec:segmentation_adaptation}), and then choosing the best configuration of hyperparameters and reporting the metrics on the test set (Section~\ref{sec:test}).
\ifthenelse{\boolean{long_version}}
{%
}
{%
	
	The Cityscapes dataset~\cite{Cordts2016} is used in all experiments, where all \textit{optimization} experiments were conducted on the validation set.
	We use four different quality measures for evaluation of the presented approach.
	These are the peak signal-to-noise ratio (PSNR)~\cite{salomon2004data}, the structural similarity (SSIM)~\cite{wang2004image}, the multi-scale structural similarity (MS-SSIM)~\cite{wang2003multi}, and the mean intersection over union (mIoU) between the reference label map and the semantic prediction of the reconstructed image produced by our own segmentation model, see Section~\ref{sec:semantic_segmentation}.
}

\begin{figure*}[t]
	\subfloat[JPEG (\SI{0.0626}{bpp})\label{fig:sample_jpeg}]{
		\includegraphics[width=0.45\textwidth]{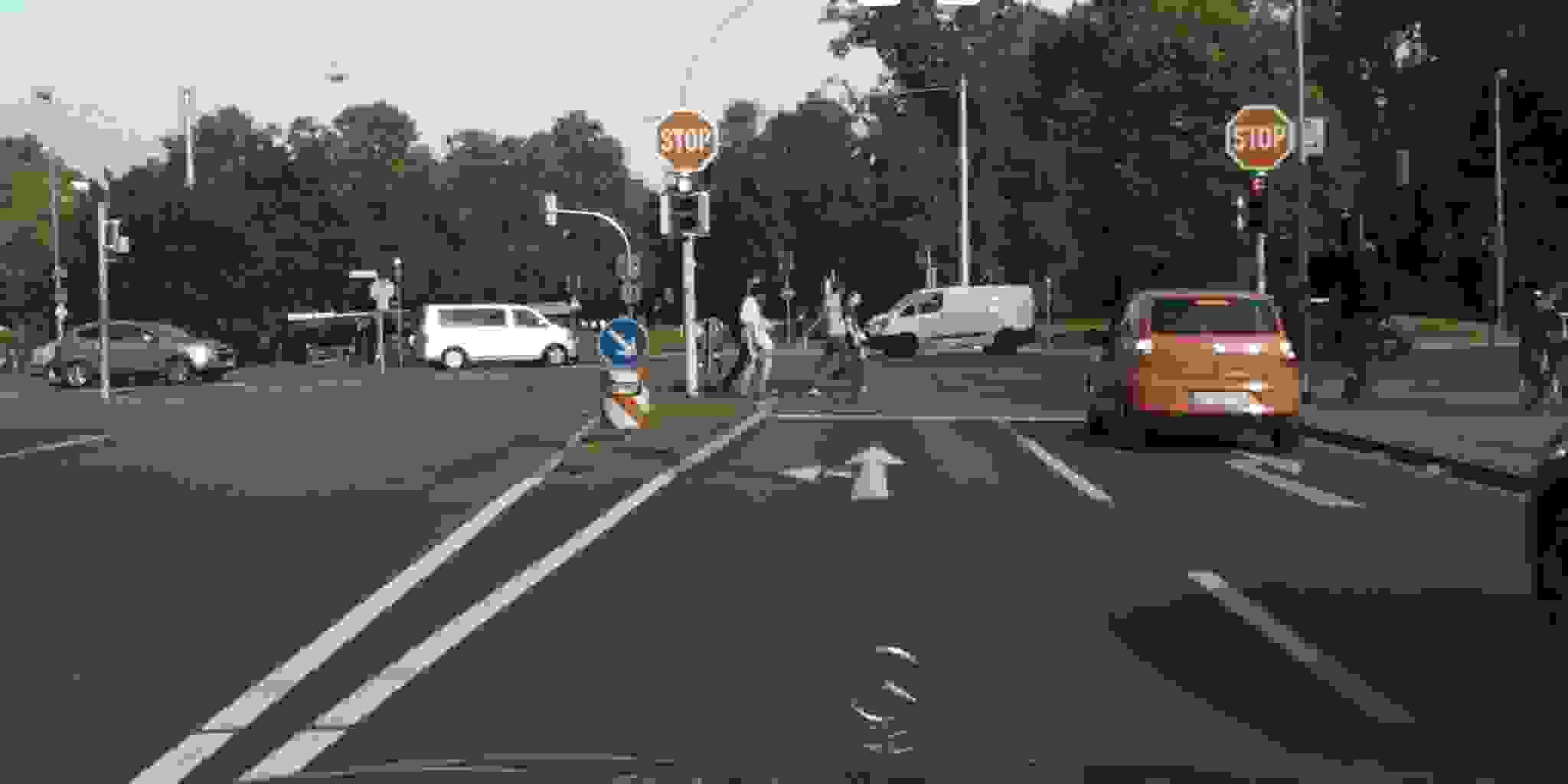}
	}
	\hfill
	\subfloat[JPEG2000 (\SI{0.0628}{bpp})\label{fig:sample_jpeg2000}]{
		\includegraphics[width=0.45\textwidth]{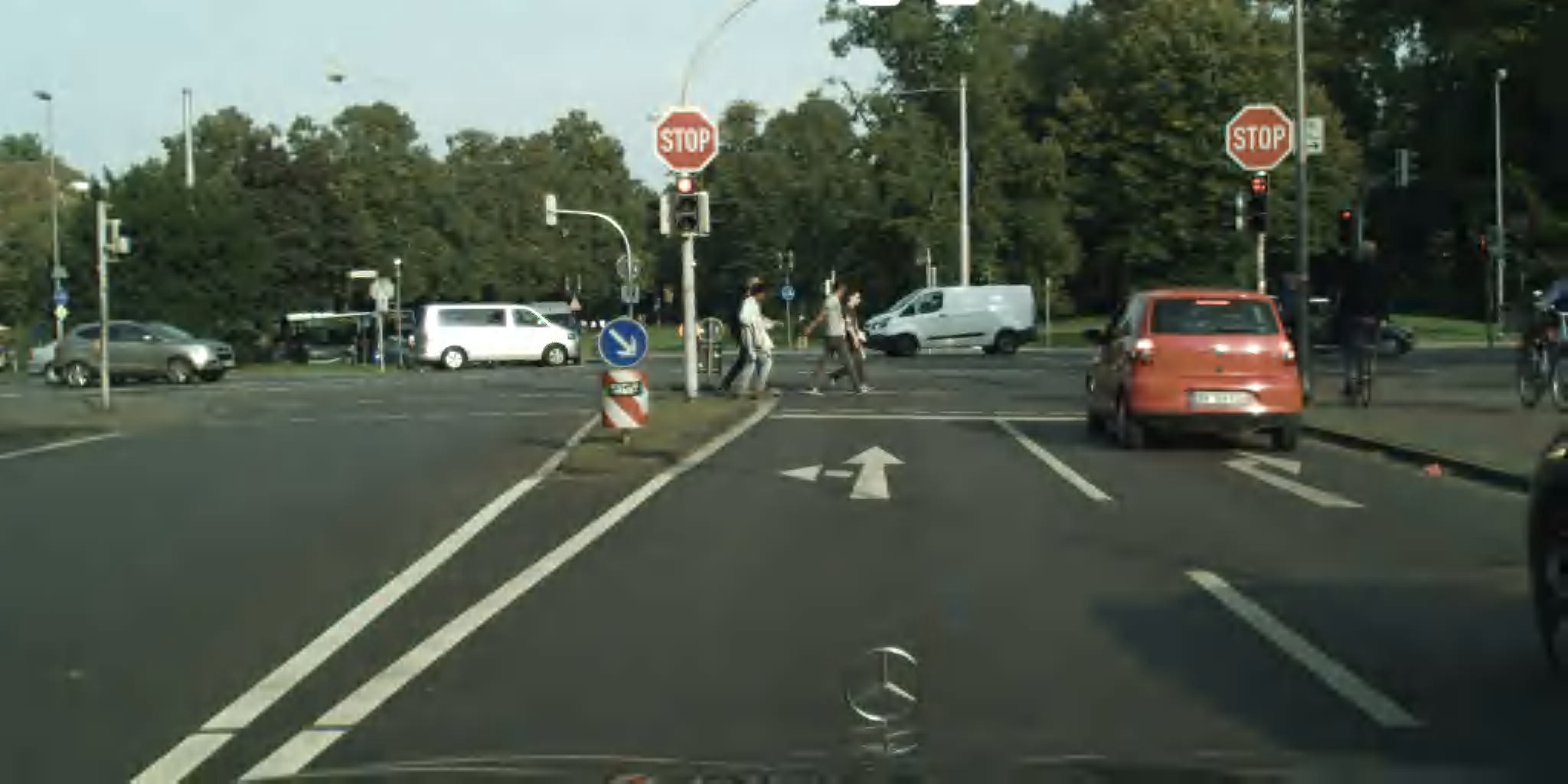}
	}
	\\[-2ex]
	\subfloat[WebP (\SI{0.0645}{bpp})\label{fig:sample_webp}]{
		\includegraphics[width=0.45\textwidth]{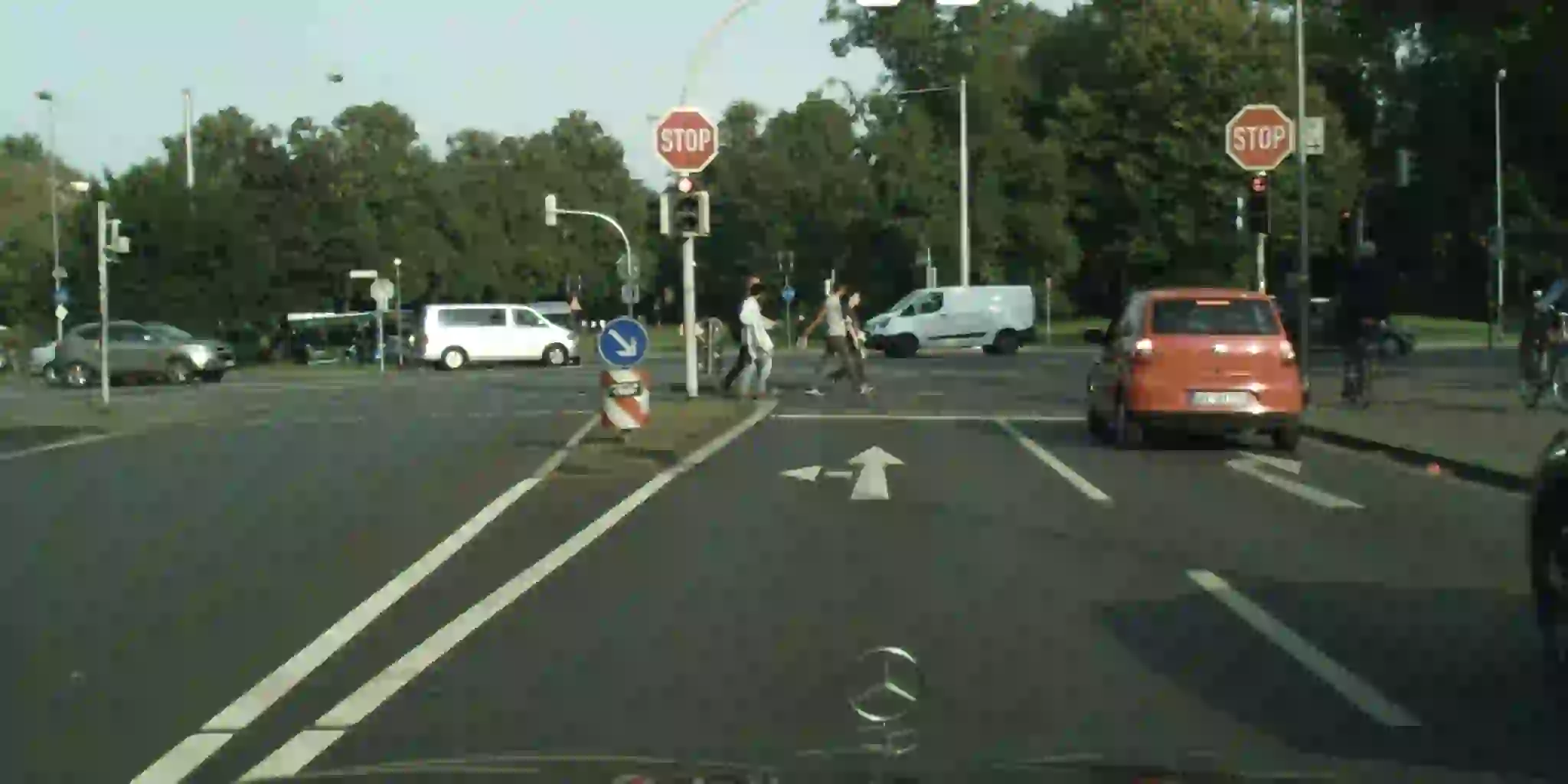}
	}
	\hfill
	\subfloat[GAN (\SI{0.0625}{bpp})\label{fig:sample_gan}]{
		\includegraphics[width=0.45\textwidth]{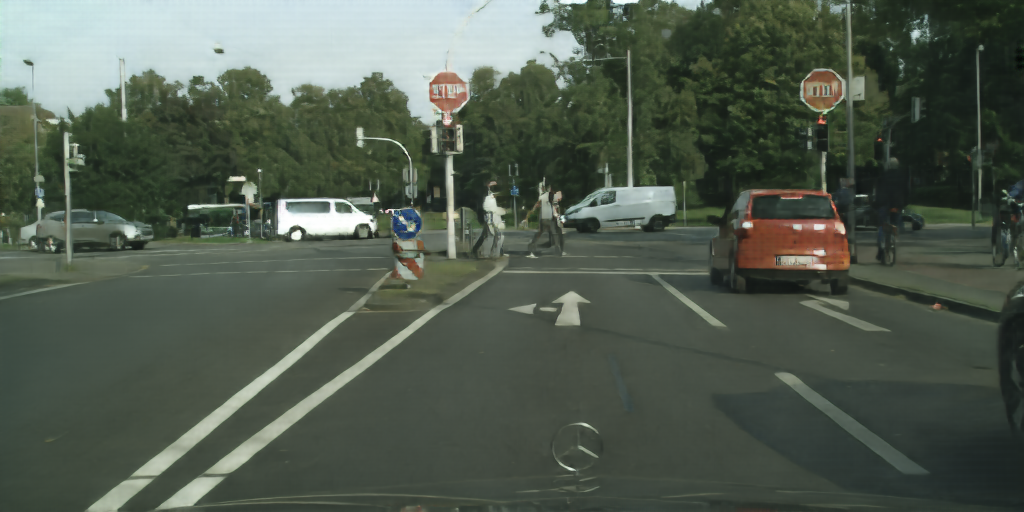}
	}
	\setlength{\belowcaptionskip}{-10pt}
	\caption{
		Samples from (a) JPEG, (b) JPEG2000, (c) WebP, and (d) GAN compression including quantization in the training process.
		The effects can be viewed best in color and on a computer screen.
	}
	\label{fig:samples}
\end{figure*}

\ifthenelse{\boolean{long_version}}
{%
	\subsection{Cityscapes Dataset}
	\label{sec:cityscapes_dataset}
	% Dataset Cityscapes
	The Cityscapes dataset~\cite{Cordts2016} is used in all experiments.
	The dataset consists of $2975$ annotated images for training, $500$ annotated images for validation and $1525$ annotated images for testing.
	All \textit{optimization} experiments in this section were conducted on the validation dataset of Cityscapes.
	To compute the mIoU on the test set, the predicted label maps have been submitted to \emph{www.cityscapes-dataset.com}.
	
	\subsection{Quality Measures}
	\label{sec:quality_measures}
	% Present PSNR, MS-SSIM, mIoU
	We use four different quality measures for evaluation of the presented approach.
	We call the first three measures distortion metrics and the latter one perception metric.
	The reference for the distortion metrics consists of the original image, which is then used for comparison to the reconstructed image.
	
	The first distortion metric is the peak signal-to-noise ratio (PSNR), which is the pixel-wise MSE between the original and the reconstructed image in decibel.
	This metric is very sensitive to variations in single image pixels and we mainly report it, because it is an important standard quality measure in the image compression domain.
	Since the goal is to restrict the image footprint to very few bits, while achieving a high reconstruction quality, we think it is more important to capture the general structure of the image than to reconstruct each pixel value exactly as it is in the reference.
	Therefore, the structural similarity (SSIM) as the second distortion metric is more suitable, because it aims at predicting the human-perceived image quality~\cite{wang2004image}.
	It compares the luminance, the contrast, and the structure of the original and the reconstructed image.
	A window function is used to compute the local statistics of the image pixels.
	The multi-scale structural similarity (MS-SSIM)~\cite{wang2003multi} is the third distortion metric and represents a successive approach which iteratively applies low-pass filters on the input images followed by downsampling by a factor of 2 in both dimensions.
	The fourth quality measure, the perception metric, differs from the former ones, as it assesses the output of the semantic segmentation network.
	The mean intersection over union (mIoU) between the reference label map and the semantic prediction of the reconstructed image is used as a metric for the preservation of the global image structure.
	We evaluate our segmentation results by using the trained stage-two model with an output stride $s=8$ (overall downsampling) and computing the mIoU between the network output and the Cityscapes ground truth.
	A similar approach was already used in~\cite{Agustsson2018} and we adopt it for the following optimizations.
}
{%
}

\subsection{Quality of Coded Images and of Semantic Segmentation on the Validation Set}
\label{sec:experiments:bitrate}
% explanation of what was done
In our GAN-based compression approach there are two parameters which can be varied to control the bitrate and quality of the generated images.
The first parameter is the number of feature maps $F$ of the bottleneck of the autoencoder.
The second parameter is the number of reconstruction levels $L$ in the quantization.
The resulting bitrate of the compression is calculated by \eqref{eq:bit_rate}.

For comparison, we used the image compression standards JPEG~\cite{jpeg}, JPEG2000~\cite{jp2}, and WebP~\cite{webp} as baselines.
The determined image size includes the file header, however, having little influence on the resulting size.
We trained each network for 50 epochs with an Adam optimizer and a learning rate of $\eta = 0.0002$ on the Cityscapes training set.
The input images are downsampled by a factor of 2 in height and width, as the limit of GPU memory is exceeded otherwise.
The metrics afterwards are computed on the validation set in the original image resolution.

\begin{figure*}[t]
	\subfloat[PSNR\label{fig:psnr}]{
		\includestandalone[width=0.48\textwidth]{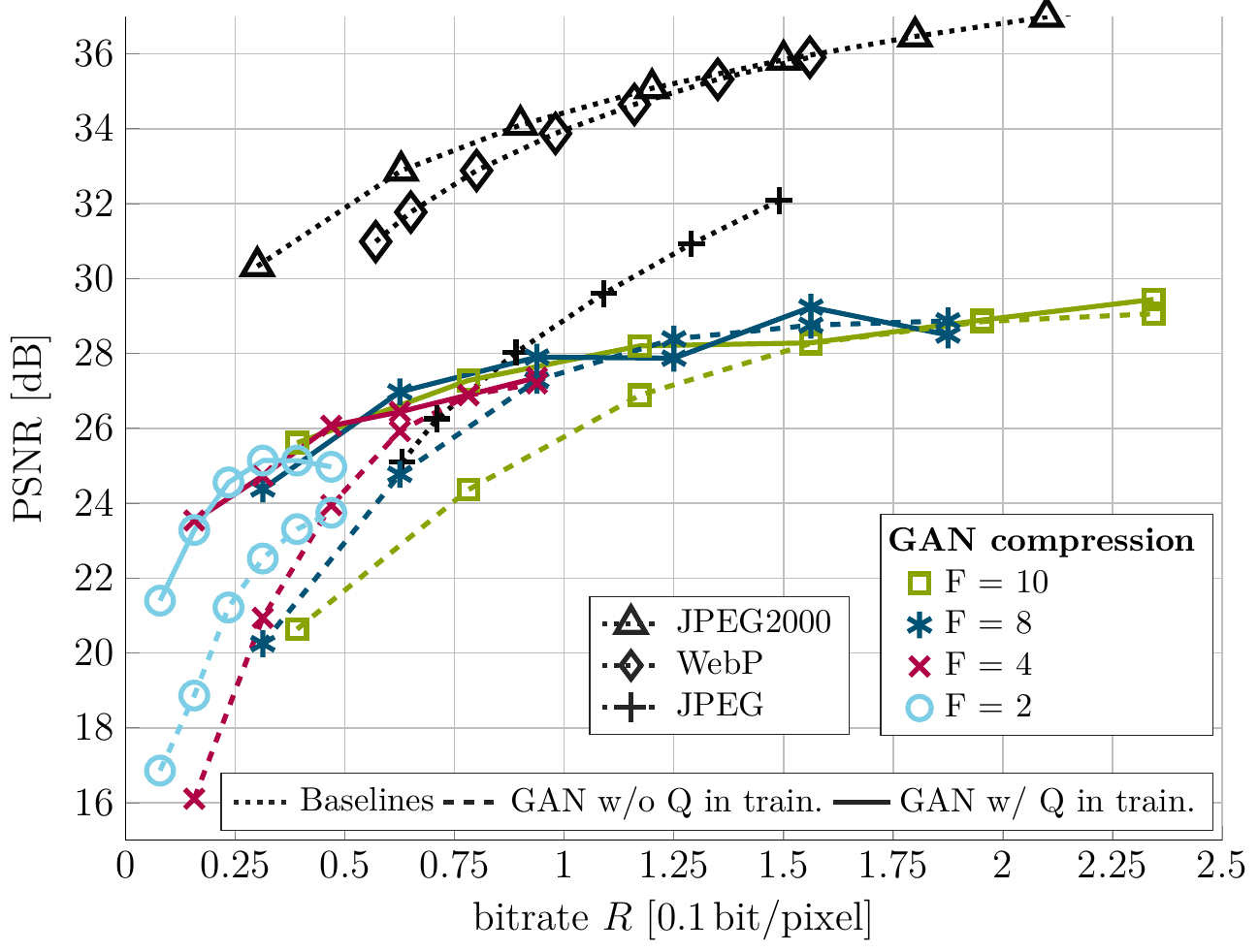}
	}
	\hfill
	\subfloat[SSIM\label{fig:ssim}]{
		\includestandalone[width=0.48\textwidth]{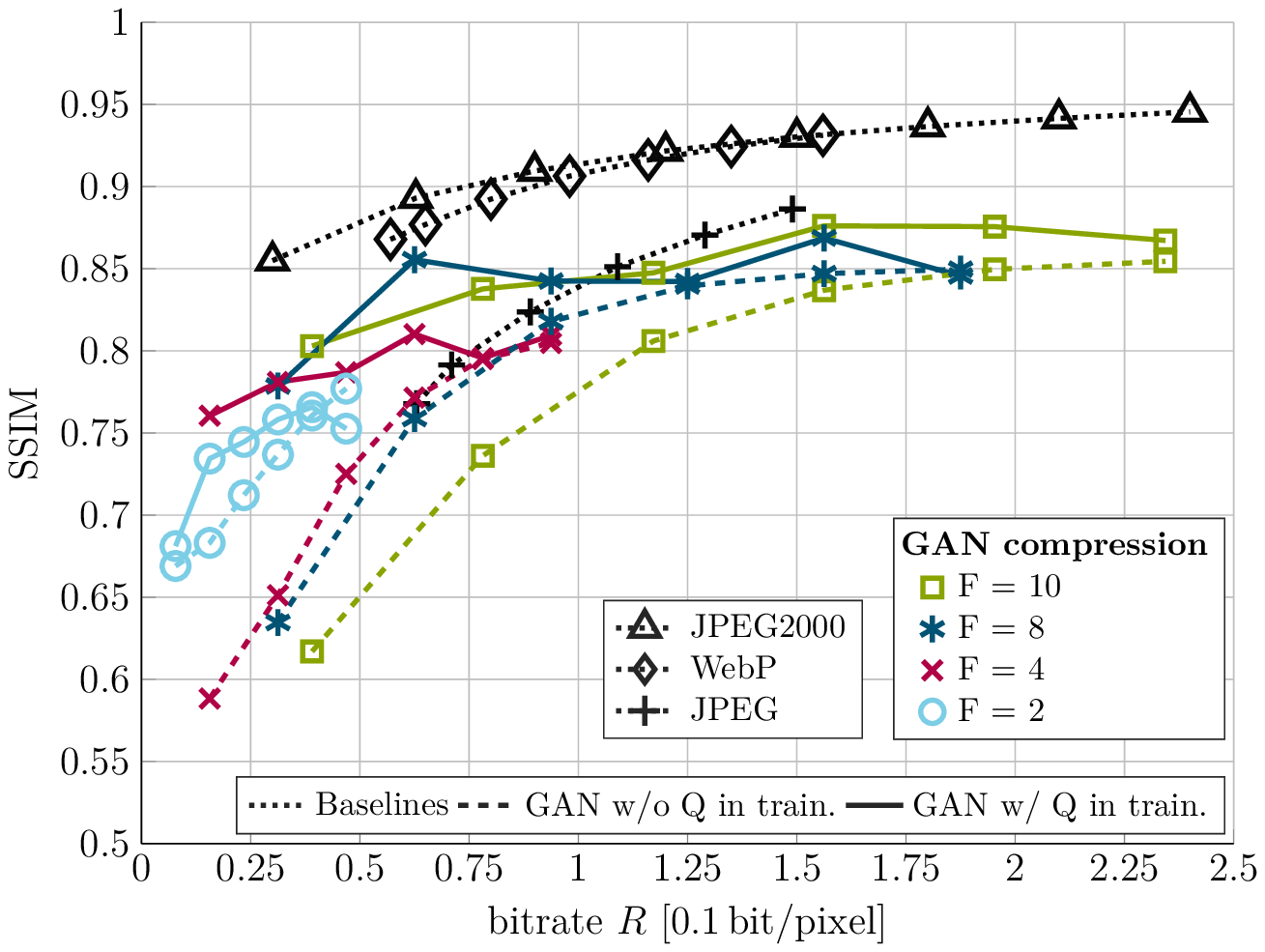}
	}
	\vspace{-0.3cm}
	\\[-1ex]
	\subfloat[MS-SSIM\label{fig:ms_ssim}]{
		\includestandalone[width=0.48\textwidth]{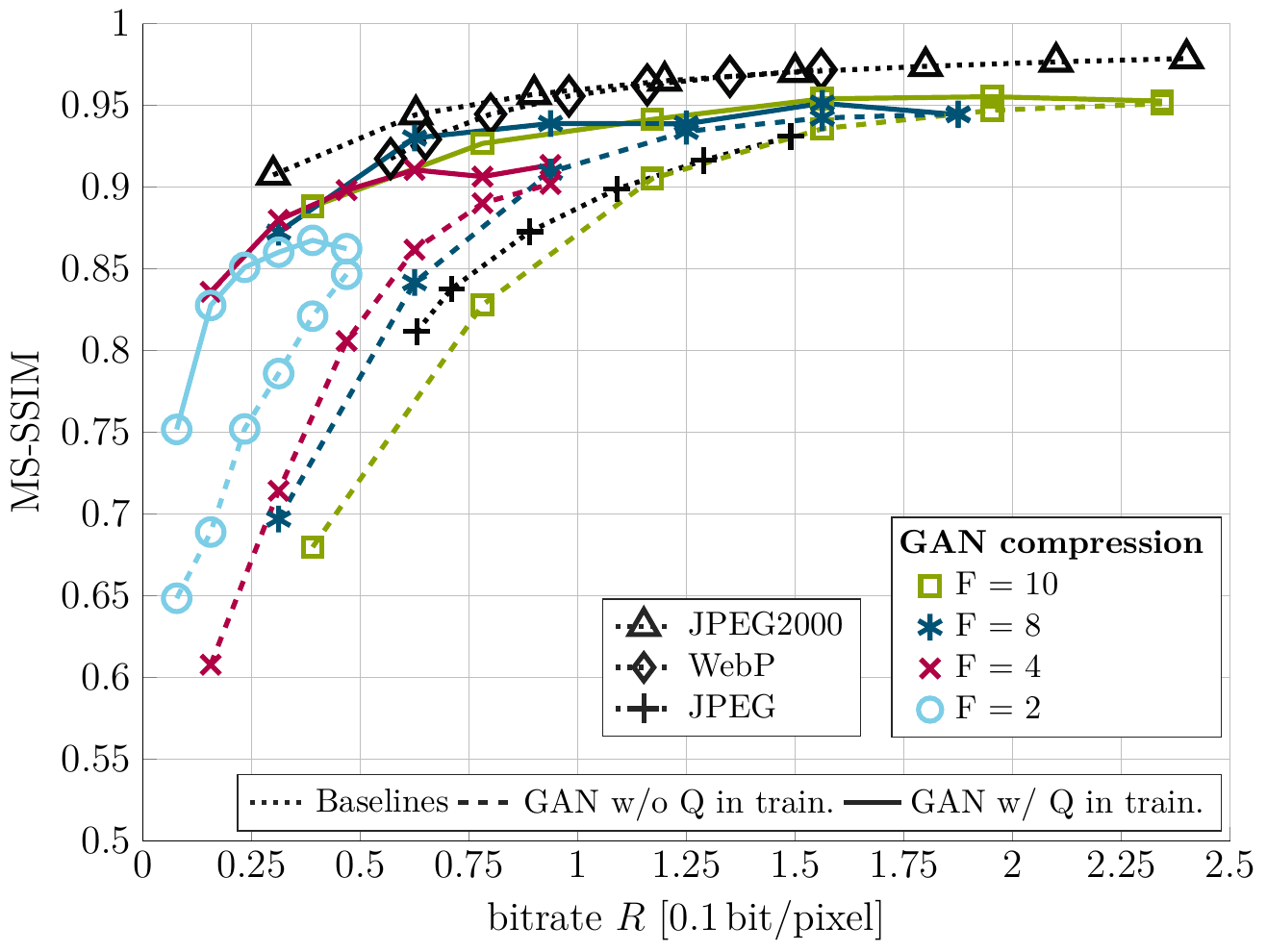}
	}
	\hfill
	\subfloat[mIoU\label{fig:miou}]{
		\includestandalone[width=0.48\textwidth]{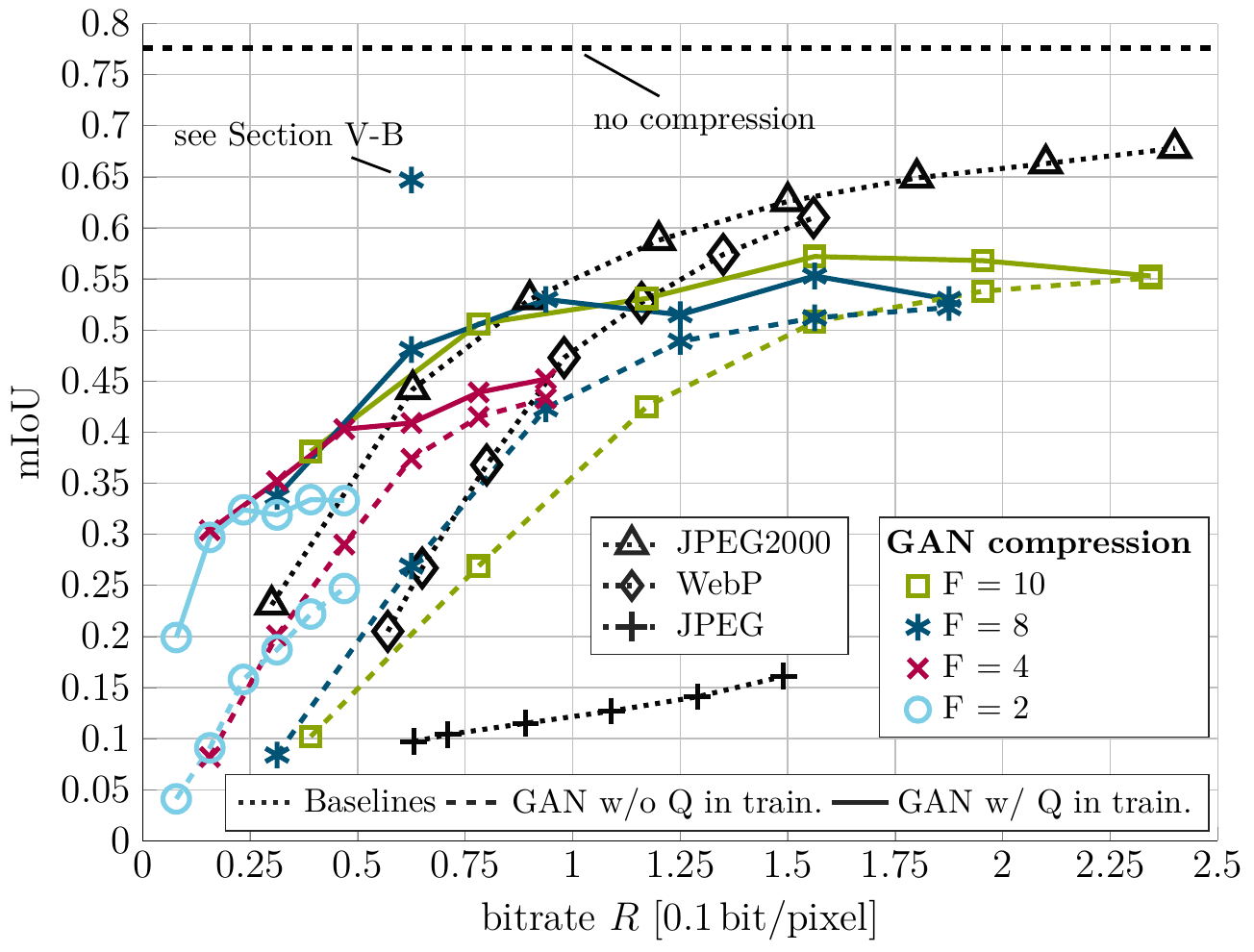}
	}
	\setlength{\belowcaptionskip}{-10pt}
	\caption{
		Evaluation of (a) PSNR, (b) SSIM, (c) MS-SSIM, and (d) mIoU for different image compression approaches on the Cityscapes \textbf{validation set}.
		The baselines are drawn in black with dotted curves.
		The GAN compression is shown without (dashed) and with (solid) quantization in the training process.
	}
	\label{fig:metrics}
\end{figure*}

Samples from the different approaches for a bitrate of \SI{0.0625}{bit} per pixel (bpp) or only slightly higher can be seen in Figure~\ref{fig:samples} (best viewed in color on a computer screen).
% block artifacts
The JPEG image has the worst perceptual quality of all shown examples and strongly reflects block artifacts, which completely remove textures and also distort the colors.
These block artifacts can also be seen in the WebP result, which may be caused by the discrete cosine transform, which is shared by JPEG and WebP.
The overall quality of JPEG2000 is best in the present comparison as it causes less distortion in small details such as traffic signs than the other approaches.
In contrast to this, when compressing by the GAN approach, the traffic signs are not well recognizable.
Nevertheless, no block artifacts are introduced by the GAN and edges are sharper than with JPEG2000 (e.g., shadow of the approaching car).

% result: 4 figures for PSNR, SSIM, MS-SSIM, and mIoU with our approaches and the baselines
% without q in training and with q in training
Figure~\ref{fig:metrics} shows the evaluations of the four quality metrics of different conditions, where the number of reconstruction levels $L$ for each pixel in the bottleneck is adapted.
In each configuration, the number of reconstruction levels is varied within \mbox{$L \in \{2,4,8,16,32,64\}$}, corresponding to $\operatorname{ld}(L)$ bit per bottleneck pixel, i.e. $\{1,2,3,4,5,6\}$ bit, respectively.
We compare models trained without quantization but using quantization during inference (dashed lines), to models trained with quantization (solid lines).
% discussion
% 1 effect of higher bitrate
As expected, with increasing bitrate all quality metrics yield higher values until they reach a plateau showing a saturation effect.

% 2 baselines vs. GAN
In terms of the PSNR (Figure~\ref{fig:psnr}) the JPEG2000 standard and WebP clearly outperform the GAN compression.
These two baselines perform quite similar, which is subjectively confirmed by Figure~\ref{fig:samples}.
JPEG2000 is slightly better, once very low bitrates are considered.
The former JPEG standard is significantly worse in PSNR and shows a steeper curve, which means that it is more sensitive to bitrate variations.
The GAN approaches without quantization in training share a similar curve, however, extended towards very low bitrates.

% 3 GAN w/ and w/o Q, GAN w/ Q for low bitrates already better than JPEG
Interestingly, Figure~\ref{fig:psnr} shows a considerably better GAN performance for the training with integrated quantization than without it, especially for low bitrates.
If the quantization is used in the training process, the encoder and decoder can adjust to the discrete reconstruction levels and yield better results.
As the bitrate increases (larger $L$), reconstruction levels get more densely populated.
In this case, the GAN approaches with or without quantization in the training show rather similar performance, as the quantization error diminishes.
It can be seen that below a bitrate of \SI{0.075}{bpp} (the intersection point with JPEG) the GANs with quantization in the training provide a higher PSNR than JPEG.

% 4 effect of F=...
Interestingly, the choice of the number of feature maps in the bottleneck $F$ only affects the PSNR of the GAN approaches without integrated quantization in training:
While lower values of $F$ yield better results at the same bitrate $R$, the bitrate itself remains the dominant quality parameter.

% 5 SSIM and MS-SSIM: In general similar behavior, but the advantage of JPEG2000 melts down
For the SSIM and MS-SSIM measures (Figures~\ref{fig:ssim}, ~\ref{fig:ms_ssim}), in general, we observe a similar behavior.
However, it is interesting to note that the GAN with quantization in training in the low bitrate regime starts to match the performance of WebP and comes very close to JPEG2000.
This can be explained by the fact, that those metrics try to evaluate images in a more perception-based way by taking into account that the human eye is more sensitive to contrasts than to color variations.
The saturation effect of all approaches is even more distinct for SSIM and MS-SSIM than for the PSNR.

% 6 mIoU
Taking a look at the overall mIoU performance of compression with subsequent semantic segmentation (Figure~\ref{fig:miou}), we find that the GAN approach is able to outperform the baselines for low bitrates ($R < 0.09$ bpp).
This aspect might be affected by the specific choice of the subsequent semantic segmentation.
Nevertheless, this interesting finding indicates that the approach favors the preservation of global structures and semantics over the reconstruction of local textures, since this is the information extracted by the segmentation.
By design, GANs prioritize enhancing the perceived image quality over the distortion metrics, because they learn the loss function themselves (in the discriminator) instead of directly optimizing for a high PSNR or SSIM.

\subsection{Quality of the Retrained Semantic Segmentation on the Validation Set}
\label{sec:segmentation_adaptation}
The second set of experiments includes now retraining of the semantic segmentation network with reconstructed images from the given GAN compression.
We call this approach retrained semantic segmentation and refer to the segmentation which is trained on original Cityscapes images as model trained with uncoded images.
When we refer to the training with GAN reconstructions, GAN encoded/decoded images are used.
We focused on a compression framework with a good tradeoff between image quality and bitrate.
The selected approach has $F=8$ feature maps in the autoencoder bottleneck and uses \SI{2}{bit} ($L=4$) for quantization of the bottleneck pixels.
The training material for the segmentation consists of the original images and the reconstructed images from Cityscapes (either mixed or one after the other, see Table~\ref*{tab:seg_results}).

\begin{table}[t]
	\vspace*{0.1cm}
	\centering
	\caption{
		Semantic segmentation mIoU results on the Cityscapes \textbf{validation set}.
		Four training strategies of the semantic segmentation (rows) are evaluated on two different validation data sets (columns):
		Original Cityscapes images (left) and reconstructed images after GAN compression (right).
		GAN coding method: $F=8, L=4$, w/ quantization in training.
	}
	\label{tab:seg_results}
	\setlength\tabcolsep{5pt}
	\begin{tabular}{cccc}
		\toprule
		& \multirow{2}{*}{\textbf{Quality Measure}} &                          \multicolumn{2}{c}{\textbf{Coding Method}}                           \\
		\cmidrule{3-4}                                           &                                           &                 \textbf{none}                 &                 \textbf{GAN}                  \\ \midrule
		&                   mIoU                    & \multirow{2}{*}{\textbf{\SI{77.6}{\percent}}} &     \multirow{2}{*}{\SI{48.1}{\percent}}      \\
		& \textbf{model trained w/ uncoded images}  &                                               &                                               \\ \midrule
		\multirow{8}{*}{\rotatebox[origin=c]{90}{\textbf{\shortstack{``Retrained''\\sem. segmentation}}}} &                   mIoU                    &     \multirow{2}{*}{\SI{66.4}{\percent}}      &     \multirow{2}{*}{\SI{63.7}{\percent}}      \\
		&  \textbf{model trained w/ GAN reconstr.}  &                                               &                                               \\
		\cmidrule{2-4}                                           &                   mIoU                    &     \multirow{2}{*}{\SI{74.0}{\percent}}      &     \multirow{2}{*}{\SI{63.4}{\percent}}      \\
		&  \textbf{model trained w/ mixed images}   &                                               &                                               \\
		\cmidrule{2-4}                                           &                   mIoU                    &     \multirow{3}{*}{\SI{71.7}{\percent}}      & \multirow{3}{*}{\textbf{\SI{64.7}{\percent}}} \\
		& \textbf{model trained w/ uncoded images,} &                                               &                                               \\
		& \textbf{then fine-tuned w/ GAN reconstr.}  &                                               &                                               \\ \bottomrule
	\end{tabular}
	\vspace*{-0.1cm}
\end{table}

The results shown in Table~\ref{tab:seg_results} are obtained from the validation set.
The first three approaches each use 120,000 iterations in the training process.
The mixture of original and reconstructed images means the union of those sets (twice the size but same number of iterations) and in case of training with uncoded images and then fine-tuning with GAN reconstructions, we train with the original images for 90,000 iterations and with the reconstructed images for another 30,000 iterations.

% discussion
As expected, when GAN compression is used, retraining of the semantic segmentation using GAN reconstructions yields significantly better mIoU scores as the segmentation purely trained with uncoded images:
The mIoU achieved by the segmentation trained with uncoded images on the validation set amounts to only \SI{48.1}{\percent}.
By combining original and reconstructed images in the training (one after the other), the best mIoU is \SI{64.7}{\percent} (blue star symbol for $F=8$, $L=4$, in Figure~\ref{fig:miou} at \SI{0.0625}{bpp}), while the performance on original images drops a bit by \SI{6}{\percent} absolute.
When fine-tuning the segmentation with reconstructions generated by JPEG2000 we obtain an mIoU of \SI{65}{\percent} being very close to the best GAN-based approach (\SI{64.7}{\percent}), but \textit{the important aspect is that from a semantic segmentation point of view GAN compression is in fact en par with the JPEG2000 image coding standard and as a fresh new technology opens the door for many more improvements.}
Remarkably, training the segmentation only with GAN reconstructions is only the second best when validating on reconstructed images.
When plotting the top GAN-based performance of \SI{64.7}{\percent} into Figure~\ref{fig:miou} (lonely blue star), \textit{we see the significant result that by using a GAN-based image compression scheme, the mIoU of a fine-tuned semantic segmentation can be improved by \SI{16.6}{\percent} absolute mIoU (\SI{64.7}{\percent}) as compared to using a segmentation model trained on uncoded images~(\SI{48.1}{\percent})}.

\subsection{Quality of the Retrained Semantic Segmentation on the Test Set}
\label{sec:test}
% Final results on the test set
The best GAN approach from the validation ($F=8$ and $L=4$) with quantization included in training, along with the original images and the JPEG2000 compression, is evaluated with the same protocol as before on the test set of the Cityscapes data set (columns in Table~\ref{tab:test}).
The three distortion metrics PSNR, SSIM, and MS-SSIM have been obtained by ourselves, since the original images are available.
The mIoU evaluation, however, requires the ground truth labels, which are not publicly available for the test set.
Therefore, we fed the segmentation network with the image compression reconstructions created from the test set to obtain the predictions and then submitted the results to \texttt{www.cityscapes-dataset.com}~\cite{Cordts2016} to obtain the mIoU scores.
% mIoU: upsampling of the prediction for the test

\begin{table}[t]
	\vspace*{0.1cm}
	\centering
	\caption{
		Final results on the Cityscapes \textbf{test set}.
		The GAN coding method is the same as in Table~\ref{tab:seg_results} ($F=8$, $L=4$, w/ quantization in training).
	}
	\label{tab:test}
	\setlength\tabcolsep{2pt}
	\begin{tabular}{cccc}
		\toprule
		  \multirow{2}{*}{\textbf{Quality Measure}}   &                                                  \multicolumn{3}{c}{\textbf{Coding Method}}                                                   \\
		               \cmidrule{2-4}                 &                 \textbf{none}                 &               \textbf{JPEG2000}               &                 \textbf{GAN}                  \\ \midrule
		                    PSNR                      &                      --                       &             \SI{34.03}{\decibel}              &             \SI{27.73}{\decibel}              \\ \midrule
		                    SSIM                      &                      --                       &                     0.908                     &                     0.871                     \\ \midrule
		                   MS-SSIM                    &                      --                       &                     0.952                     &                     0.938                     \\ \midrule
		                    mIoU                      & \multirow{2}{*}{\textbf{\SI{75.3}{\percent}}} &     \multirow{2}{*}{\SI{46.0}{\percent}}      &     \multirow{2}{*}{\SI{50.4}{\percent}}      \\
		  \textbf{model trained w/ uncoded images}    &                                               &                                               &                                               \\ \midrule
		                    mIoU                      &     \multirow{3}{*}{\SI{70.83}{\percent}}     & \multirow{3}{*}{\textbf{\SI{65.0}{\percent}}} &     \multirow{3}{*}{\SI{59.69}{\percent}}     \\
		  \textbf{model trained w/ uncoded images,}   &                                               &                                               &                                               \\
		\textbf{then fine-tuned w/ JPEG2000 reconstr.} &                                               &                                               &                                               \\ \midrule
		                    mIoU                      &     \multirow{3}{*}{\SI{66.3}{\percent}}      &     \multirow{3}{*}{\SI{57.6}{\percent}}      & \multirow{3}{*}{\textbf{\SI{63.0}{\percent}}} \\
		  \textbf{model trained w/ uncoded images,}   &                                               &                                               &                                               \\
		  \textbf{then fine-tuned w/ GAN reconstr.}    &                                               &                                               &                                               \\ \bottomrule
	\end{tabular}
	\vspace*{-0.1cm}
\end{table}

The final results on the test set can be found in Table~\ref{tab:test}.
They reflect the same effects as discovered on the validation set.
The distortion metrics yield higher values for JPEG2000 than for GAN compression, with the margin shrinking for the structural similarity metrics.
Concerning mIoU results, no coding and both coding methods are best when performing fine-tuning with respective reconstructions.
\textit{Interestingly, the GAN-based coding yields even better mIoU performance than JPEG2000, if the semantic segmentation is simply trained on uncoded images, although JPEG2000 delivers the better PSNR.
Again, we see both JPEG2000 and GAN compression to perform comparably well in the context of semantic segmentation, with GAN compression having a high potential for further improvement.}

\section{Conclusions}
\label{sec:conclusions}
In this paper, we compared several image compression standards to a GAN-based image compression, making use of an autoencoder as the generator part of the adversarial network.
Our evaluation concentrated on typical distortion metrics like PSNR and SSIM, but also on the performance of a subsequent perception function in form of a semantic segmentation based on \texttt{DeepLabv3}, which is our perception metric.
As expected, our experiments showed that a higher bitrate leads to better performance in all quality measures.
In terms of the traditional metrics, the GAN-based compression is more or less outperformed by the standards.
Concerning segmentation performance, however, the GAN yields a mean intersection over union (mIoU) comparable to JPEG2000, particularly at low bitrates.
The performance gain that can be obtained by retraining the semantic segmentation with GAN reconstructions has been shown to be enormous.
If no retraining is performed, e.g., to keep modularity of the compression and semantic segmentation blocks, the GAN-based compression exceeds JPEG2000 by more than \SI{4}{\percent} absolute in mIoU.

% important conclusion --> not to short
The performance of subsequent functions is a major point of interest in automotive applications.
As could be shown, traditional metrics may not be well-suited for this purpose, as their compression affects the segmentation result in a negative way if the compression ratio becomes too high.
Therefore, new algorithms are required for bitrate-efficient distributed perception.
Compression approaches using an adversarial loss have been shown to have the potential to yield high semantic segmentation performance even if they do not reconstruct as close to the original pixel values as conventional image compression methods do.

\section{Acknowledgment}
The authors gratefully acknowledge support of this work by Volkswagen Group Research, Wolfsburg, Germany.
\ifthenelse{\boolean{long_version}}
{%
	\addtolength{\textheight}{-3.5cm}   % This command serves to balance the column lengths
}
{%
%	\addtolength{\textheight}{-3.5cm}
}

%\section*{Appendix}
%\label{sec:appendix}

%\section*{ACKNOWLEDGMENT}

\bibliographystyle{IEEEtran}
%\bibliography{IEEEabrv,Q:/Documents/Publikationen/Bibliographie/bib_computer_vision}
\bibliography{IEEEabrv,bib_computer_vision}

\begin{thebibliography}{10}
\providecommand{\url}[1]{#1}
\csname url@rmstyle\endcsname
\providecommand{\newblock}{\relax}
\providecommand{\bibinfo}[2]{#2}
\providecommand\BIBentrySTDinterwordspacing{\spaceskip=0pt\relax}
\providecommand\BIBentryALTinterwordstretchfactor{4}
\providecommand\BIBentryALTinterwordspacing{\spaceskip=\fontdimen2\font plus
\BIBentryALTinterwordstretchfactor\fontdimen3\font minus
  \fontdimen4\font\relax}
\providecommand\BIBforeignlanguage[2]{{%
\expandafter\ifx\csname l@#1\endcsname\relax
\typeout{** WARNING: IEEEtran.bst: No hyphenation pattern has been}%
\typeout{** loaded for the language `#1'. Using the pattern for}%
\typeout{** the default language instead.}%
\else
\language=\csname l@#1\endcsname
\fi
#2}}

\bibitem{jpeg}
ISO/ITU, ``{Information Technology -- Digital Compression and Coding of
  Continuous-Tone Still Images: Requirements and Guidelines},'' 1992,
  {published by ISO as 10918-1 and ITU as T.81}.

\bibitem{jp2}
------, ``{Information technology -- JPEG 2000 image coding system: Core coding
  system},'' 2016, {published by ISO as 15444-1 and ITU as T.800}.

\bibitem{webp}
{Google Developers WebP}, ``{WebP image format},''
  \url{https://developers.google.com/speed/webp/}, 2015, accessed: 2018-07-24.

\bibitem{salomon2004data}
D.~Salomon, \emph{Data Compression: The Complete Reference}.\hskip 1em plus
  0.5em minus 0.4em\relax Springer Science \& Business Media, 2004.

\bibitem{wang2004image}
Z.~Wang, A.~C. Bovik, H.~R. Sheikh, and E.~P. Simoncelli, ``{Image Quality
  Assessment: From Error Visibility to Structural Similarity},'' \emph{IEEE
  Transactions on Image Processing}, vol.~13, no.~4, pp. 600--612, Apr. 2004.

\bibitem{wang2003multi}
Z.~Wang, E.~Simoncelli, A.~Bovik, \emph{et~al.}, ``{Multi-Scale Structural
  Similarity for Image Quality Assessment},'' in \emph{Proc. of ACSSC}, Pacific
  Grove, CA, USA, Nov. 2003, pp. 1398--1402.

\bibitem{Isola2016}
P.~Isola, J.-Y. Zhu, T.~Zhou, and A.~A. Efros, ``{Image-to-Image Translation
  with Conditional Adversarial Networks},'' in \emph{Proc. of CVPR}, Honolulu,
  HI, USA, July 2017, pp. 1125--1134.

\bibitem{wang2018high}
T.-C. Wang, M.-Y. Liu, J.-Y. Zhu, A.~Tao, J.~Kautz, and B.~Catanzaro,
  ``{High-Resolution Image Synthesis and Semantic Manipulation with Conditional
  GANs},'' in \emph{Proc. of CVPR}, Salt Lake City, UT, USA, June 2018, pp.
  8798--8807.

\bibitem{Blau2017}
Y.~Blau and T.~Michaeli, ``{The Perception-Distortion Tradeoff},'' in
  \emph{Proc. of CVPR}, Salt Lake City, UT, USA, June 2018, pp. 6228--6237.

\bibitem{Goodfellow2014}
I.~J. Goodfellow, J.~Pouget-Abadie, M.~Mirza, B.~Xu, D.~Warde-Farley, S.~Ozair,
  A.~Courville, and Y.~Bengio, ``{Generative Adversarial Nets},'' in
  \emph{Proc. of NIPS}, Montr{\'{e}}al, Canada, Dec. 2014, pp. 2672--2680.

\bibitem{Agustsson2018}
E.~Agustsson, M.~Tschannen, F.~Mentzer, R.~Timofte, and L.~V. Gool,
  ``{Generative Adversarial Networks for Extreme Learned Image Compression},''
  \emph{arXiv}, Apr. 2018.

\bibitem{Mirza2014}
M.~Mirza and S.~Osindero, ``{Conditional Generative Adversarial Nets},''
  \emph{arXiv}, Nov. 2014.

\bibitem{Radford2015}
A.~Radford, L.~Metz, and S.~Chintala, ``{Unsupervised Representation Learning
  with Deep Convolutional Generative Adversarial Networks},'' in \emph{Proc. of
  ICLR}, San Juan, Puerto Rico, May 2016, pp. 1--16.

\bibitem{Mao2016}
X.~Mao, Q.~Li, H.~Xie, R.~Y.~K. Lau, Z.~Wang, and S.~P. Smolley, ``{Least
  Squares Generative Adversarial Networks},'' in \emph{Proc. of ICCV}, Venice,
  Italy, Oct. 2017, pp. 2794--2802.

\bibitem{Zhao2016}
J.~Zhao, M.~Mathieu, and Y.~LeCun, ``{Energy-based Generative Adversarial
  Network},'' in \emph{Proc. of ICLR}, Toulon, France, Apr. 2017, pp. 1--17.

\bibitem{Berthelot2017}
D.~Berthelot, T.~Schumm, and L.~Metz, ``{BEGAN: Boundary Equilibrium Generative
  Adversarial Networks},'' \emph{arXiv}, Mar. 2017.

\bibitem{Arjovsky2017}
M.~Arjovsky, S.~Chintala, and L.~Bottou, ``{Wasserstein GAN},'' in \emph{Proc.
  of ICML}, Sydney, Australia, Aug. 2017, pp. 214--223.

\bibitem{Kurach2018}
K.~Kurach, M.~Lucic, X.~Zhai, M.~Michalski, and S.~Gelly, ``{The GAN Landscape:
  Losses, Architectures, Regularization, and Normalization},'' \emph{arXiv},
  July 2018.

\bibitem{Makhzani2015}
A.~Makhzani, J.~Shlens, N.~Jaitly, and I.~Goodfellow, ``{Adversarial
  Autoencoders},'' in \emph{Proc. of ICLR}, San Juan, Puerto Rico, May 2016,
  pp. 1--16.

\bibitem{Luc2016}
P.~Luc, C.~Couprie, S.~Chintala, and J.~Verbeek, ``{Semantic Segmentation using
  Adversarial Networks},'' in \emph{NIPS Workshop on Adversarial Training},
  Barcelona, Spain, Dec. 2016, pp. 1--12.

\bibitem{Belagiannis2018}
V.~Belagiannis, A.~Farshad, and F.~Galasso, ``{Adversarial Network
  Compression},'' \emph{arXiv}, Mar. 2018.

\bibitem{Xiao2018}
C.~Xiao, B.~Li, J.-Y. Zhu, W.~He, M.~Liu, and D.~Song, ``{Generating
  Adversarial Examples with Adversarial Networks},'' \emph{arXiv}, Jan. 2018.

\bibitem{Samangouei2018}
P.~Samangouei, M.~Kabkab, and R.~Chellappa, ``{Defense-GAN: Protecting
  Classifiers Against Adversarial Attacks Using Generative Models},'' in
  \emph{Proc. of ICLR}, Vancouver, Canada, Apr. 2018, pp. 1--17.

\bibitem{Ilyas2017}
A.~Ilyas, A.~Jalal, E.~Asteri, C.~Daskalakis, and A.~G. Dimakis, ``{The Robust
  Manifold Defense: Adversarial Training using Generative Models},''
  \emph{arXiv}, Dec. 2017.

\bibitem{Ho2016}
J.~Ho and S.~Ermon, ``{Generative Adversarial Imitation Learning},'' in
  \emph{Proc. of NIPS}, Barcelona, Spain, Dec. 2016, pp. 4565--4573.

\bibitem{Shrivastava2016}
A.~Shrivastava, T.~Pfister, O.~Tuzel, J.~Susskind, W.~Wang, and R.~Webb,
  ``{Learning from Simulated and Unsupervised Images through Adversarial
  Training},'' in \emph{Proc. of CVPR}, Honulu, HI, USA, July 2017, pp.
  2107--2116.

\bibitem{Zhu2017}
J.-Y. Zhu, T.~Park, P.~Isola, and A.~A. Efros, ``{Unpaired Image-to-Image
  Translation using Cycle-Consistent Adversarial Networks},'' in \emph{Proc. of
  ICCV}, Venice, Italy, Oct. 2017, pp. 2223--2232.

\bibitem{Li2018}
P.~Li, X.~Liang, D.~Jia, and E.~P. Xing, ``{Semantic-aware Grad-GAN for
  Virtual-to-Real Urban Scene Adaption},'' in \emph{Proc. of BMVC}, Newcastle,
  England, Sept. 2018, pp. 1--13.

\bibitem{Sankaranarayanan2017}
S.~Sankaranarayanan, Y.~Balaji, A.~Jain, S.~N. Lim, and R.~Chellappa,
  ``{Learning from Synthetic Data: Addressing Domain Shift for Semantic
  Segmentation},'' in \emph{Proc. of CVPR}, Salt Lake City, UT, USA, June 2018,
  pp. 3752--3761.

\bibitem{Theis2017}
L.~Theis, W.~Shi, A.~Cunningham, and F.~Huszár, ``{Lossy Image Compression
  with Compressive Autoencoders},'' in \emph{Proc. of ICLR}, Toulon, France,
  Apr. 2017, pp. 1--19.

\bibitem{Balle2016}
J.~Ballé, V.~Laparra, and E.~P. Simoncelli, ``{End-to-end Optimized Image
  Compression},'' in \emph{Proc. of ICLR}, Toulon, France, Apr. 2017, pp.
  1--27.

\bibitem{Rippel2017}
O.~Rippel and L.~Bourdev, ``{Real-Time Adaptive Image Compression},'' in
  \emph{Proc. of ICML}, Sydney, Australia, Aug. 2017, pp. 2922--2930.

\bibitem{Toderici2015}
G.~Toderici, S.~M. O'Malley, S.~J. Hwang, D.~Vincent, D.~Minnen, S.~Baluja,
  M.~Covell, and R.~Sukthankar, ``{Variable Rate Image Compression with
  Recurrent Neural Networks},'' \emph{arXiv}, Nov. 2015.

\bibitem{Santurkar2017}
S.~Santurkar, D.~Budden, and N.~Shavit, ``{Generative Compression},''
  \emph{arXiv}, Mar. 2017.

\bibitem{Long2015}
J.~Long, E.~Shelhamer, and T.~Darrell, ``{Fully Convolutional Networks for
  Semantic Segmentation},'' in \emph{Proc. of CVPR}, Boston, MA, USA, June
  2015, pp. 3431--3440.

\bibitem{Russakovsky2015}
O.~Russakovsky, J.~Deng, H.~Su, J.~Krause, S.~Satheesh, S.~Ma, Z.~Huang,
  A.~Karpathy, A.~Khosla, M.~Bernstein, A.~C. Berg, and L.~Fei-Fei, ``{ImageNet
  Large Scale Visual Recognition Challenge},'' \emph{International Journal of
  Computer Vision (IJCV)}, vol. 115, no.~3, pp. 211--252, Dec. 2015.

\bibitem{He2016}
K.~He, X.~Zhang, S.~Ren, and J.~Sun, ``{Deep Residual Learning for Image
  Recognition},'' in \emph{Proc. of CVPR}, Las Vegas, NV, USA, June 2016, pp.
  770--778.

\bibitem{Zhao2016a}
H.~Zhao, J.~Shi, X.~Qi, X.~Wang, and J.~Jia, ``{Pyramid Scene Parsing
  Network},'' in \emph{Proc. of CVPR}, Honulu, HI, USA, July 2017, pp.
  2881--2890.

\bibitem{Chen2017}
L.~C. Chen, G.~Papandreou, F.~Schroff, and H.~Adam, ``{Rethinking Atrous
  Convolution for Semantic Image Segmentation},'' \emph{arXiv}, June 2017.

\bibitem{Wu2016}
Z.~Wu, C.~Shen, and A.~van~den Hengel, ``{Wider or Deeper: Revisiting the
  ResNet Model for Visual Recognition},'' \emph{arXiv}, Nov. 2016.

\bibitem{Chen2018a}
L.~Chen, Y.~Zhu, G.~Papandreou, F.~Schroff, and H.~Adam, ``{Encoder-Decoder
  with Atrous Separable Convolution for Semantic Image Segmentation},'' in
  \emph{Proc. of ECCV}, Munich, Germany, Sept. 2018, pp. 801--818.

\bibitem{RotaBulo2018}
S.~Rota~Bulò, L.~Porzi, and P.~Kontschieder, ``{In-Place Activated BatchNorm
  for Memory-Optimized Training of DNNs},'' in \emph{Proc. of CVPR}, Salt Lake
  City, UT, USA, June 2018, pp. 5639--5647.

\bibitem{Yu2016}
F.~Yu and V.~Koltun, ``{Multi-Scale Context Aggregation by Dilated
  Convolutions},'' in \emph{Proc. of ICLR}, San Juan, Puerto Rico, May 2016,
  pp. 1--13.

\bibitem{Chen2018}
L.~C. Chen, G.~Papandreou, I.~Kokkinos, K.~Murphy, and A.~L. Yuille,
  ``{DeepLab: Semantic Image Segmentation with Deep Convolutional Nets, Atrous
  Convolution, and Fully Connected CRFs},'' \emph{IEEE Transactions on Pattern
  Analysis and Machine Intelligence (TPAMI)}, vol.~40, no.~4, pp. 834--848,
  Apr. 2018.

\bibitem{Ronneberger2015}
O.~Ronneberger, P.~Fischer, and T.~Brox, ``{U-Net: Convolutional Networks for
  Biomedical Image Segmentation},'' in \emph{Proc. of MICCAI}, Munich, Germany,
  Oct. 2015, pp. 234--241.

\bibitem{Ioffe2015}
S.~Ioffe and C.~Szegedy, ``{Batch Normalization: Accelerating Deep Network
  Training by Reducing Internal Covariate Shift},'' in \emph{Proc. of ICML},
  Lille, France, July 2015, pp. 448--456.

\bibitem{Sandler2018}
M.~Sandler, A.~G. Howard, M.~Zhu, A.~Zhmoginov, and L.~Chen, ``{MobileNetV2{:}
  Inverted Residuals and Linear Bottlenecks},'' in \emph{Proc. of CVPR}, Salt
  Lake City, UT, USA, June 2018, pp. 4510--4520.

\bibitem{Romera2018}
E.~Romera, J.~M. Álvarez, L.~M. Bergasa, and R.~Arroyo, ``{ERFNet: Efficient
  Residual Factorized Conv\-Net for Real-Time Semantic Segmentation},''
  \emph{IEEE Transactions on Intelligent Transportation Systems (ITS)},
  vol.~19, no.~1, pp. 263--272, Jan. 2018.

\bibitem{salimans2016improved}
T.~Salimans, I.~Goodfellow, W.~Zaremba, V.~Cheung, A.~Radford, and X.~Chen,
  ``{Improved techniques for training GANs},'' in \emph{Proc. of NIPS},
  Barcelona, Spain, Dec. 2016, pp. 2234--2242.

\bibitem{Heusel2017}
M.~Heusel, H.~Ramsauer, T.~Unterthiner, B.~Nessler, and S.~Hochreiter, ``{GANs
  Trained by a Two Time-Scale Update Rule Converge to a Local Nash
  Equilibrium},'' in \emph{Proc. of NIPS}, Los Angeles, CA, USA, Dec. 2017, pp.
  6626--6637.

\bibitem{borji2018pros}
A.~Borji, ``{Pros and Cons of GAN Evaluation Measures},'' \emph{arXiv}, Feb.
  2018.

\bibitem{Everingham2015}
M.~Everingham, S.~M. Eslami, L.~Gool, C.~K. Williams, J.~Winn, and
  A.~Zisserman, ``{The Pascal Visual Object Classes Challenge: A
  Retrospective},'' \emph{International Journal of Computer Vision (IJCV)},
  vol. 111, no.~1, pp. 98--136, Jan. 2015.

\bibitem{JTan}
J.~Tan, ``{TensorFlow Implementation of Generative Adversarial Networks for
  Extreme Learned Image Compression},''
  \url{https://github.com/Justin-Tan/generative-compression}, 2018, accessed:
  2018-07-24.

\bibitem{bengio-arxiv13-condcomp}
Y.~Bengio, N.~L{\'{e}}onard, and A.~Courville, ``{Estimating or Propagating
  Gradients Through Stochastic Neurons for Conditional Computation},''
  \emph{arXiv}, Aug. 2013.

\bibitem{Abadi2016}
M.~Abadi \emph{et~al.}, ``{TensorFlow: Large-Scale Machine Learning on
  Heterogeneous Distributed Systems},'' \emph{arXiv}, Mar. 2016.

\bibitem{Cordts2016}
M.~Cordts, M.~Omran, S.~Ramos, T.~Rehfeld, M.~Enzweiler, R.~Benenson,
  U.~Franke, S.~Roth, and B.~Schiele, ``{The Cityscapes Dataset for Semantic
  Urban Scene Understanding},'' in \emph{Proc. of CVPR}, Las Vegas, NV, USA,
  June 2016, pp. 3213--3223.

\end{thebibliography}

\end{document}